\documentclass[10pt,a4paper]{article}
\usepackage[numbers]{natbib}
\usepackage{amsmath,amssymb,amsfonts}
\usepackage{graphicx}
\usepackage{textcomp}
\usepackage{algorithm}
\usepackage{algorithmicx}
\usepackage{algpseudocode}
\usepackage{threeparttable}
\usepackage[caption=false]{subfig}
\usepackage{booktabs}
\usepackage{url}
\usepackage{hyperref}
\usepackage{multirow}
\usepackage{tabularx}
\usepackage{authblk}
\usepackage{fullpage}
\usepackage{microtype}

\title{Computer Vision based inspection on post-earthquake with UAV synthetic dataset}

\author[1,2]{Mateusz \.Zarski}
\author[1,2]{Bartosz Wójcik}
\author[2]{Jaros\l{}aw A. Miszczak}
\author[3]{Bartlomiej Blachowski}
\author[3]{Mariusz Ostrowski}

\affil[1]{Faculty of Civil Engineering, Silesian University of Technology,\newline Akademicka 5, 44-100 Gliwice, Poland}
\affil[2]{Institute of Theoretical and Applied Informatics, Polish Academy of Sciences,\newline Ba\l{}tycka 5, 44-100 Gliwice, Poland}

\affil[3]{Institute of Fundamental Technological Research, Polish Academy of Sciences, Pawińskiego 5B, 02-106 Warsaw, Poland}

\begin{document}

%
%

%

\maketitle

\begin{abstract}
The area affected by the earthquake is vast and often difficult to entirely cover, and the earthquake itself is a sudden event that causes multiple defects simultaneously, that cannot be effectively traced using traditional, manual methods. This article presents an innovative approach to the problem of detecting damage after sudden events by using an interconnected set of deep machine learning models organized in a single pipeline and allowing for easy modification and swapping models seamlessly. Models in the pipeline were trained with a synthetic dataset and were adapted to be further evaluated and used with unmanned aerial vehicles (UAVs) in real-world conditions.
Thanks to the methods presented in the article, it is possible to obtain high accuracy in detecting buildings defects, segmenting constructions into their components and estimating their technical condition based on a single drone flight.
\end{abstract}


\section{Introduction}\label{sec:intro}

Earthquakes are sudden and violent disasters that cover huge areas of land in a very short period of time. They have been known to mankind since ancient times and invariably pose one of the most serious threats to the lives of people concentrated in large cities. The scale of their destructive power can be seen in the number of nearly two million earthquake victims in the 20th century alone~\cite{xxdeathtoll}, or in the most devastating events, which could claim up to nearly a million lives~\cite{biggestdeathtoll}. At the same time, the map of seismically active areas largely overlaps with densely populated areas, particularly in North America, Europe and Asia~\cite{quakes_desc}, which focuses researchers on this type of hazard and methods of its mitigation.

Studies conducted to date have assessed the effects of earthquakes both in terms of the impact on housing and infrastructure, and the performance of public services in repairing damage or improving traffic flow in the affected area~\cite{quake2, quake3}. These works have led to concepts of cities in which such events will no longer have a critical impact on the lives of residents, but with the cost of monitoring the condition of structures even after seemingly harmless, small earthquakes to take corrective action immediately after damage occurs~\cite{quake4}. This, however, requires the use of modern methods of construction monitoring to reduce the labor intensity of the entire process, without which the end goal is impossible to achieve.

In this paper, we present our step towards building autonomous systems that can bring this goal closer. Using a synthetic dataset containing models of earthquake-damaged buildings observed from unmanned aerial vehicle (UAV) like perspective, we created a robust, yet easily modifiable pipeline featuring multiple machine learning models that can be applied in real-life scenarios. The models we have trained allow us to detect close objects, segment them into their components, and finally detect their defects and evaluate their condition. In our work, we also described the specifics of working with a synthetic dataset and the possibilities for extending our solution that could improve its accuracy.

The rest of this paper is organized as follows. In Section~\ref{sec:related_works} we present works related to the main topics of the article, including earthquake-induced structural defects, defect detection and the use of UAVs along with synthetic dataset. In Section~\ref{sec:our_approach} we describe our approach starting with dataset description. Then we cover in detail each task of our approach together with our solution and develop our final pipeline. Section~\ref{sec:discussion} provides discussion of our solution. in which we suggest ways to improve it. Lastly, in Section~\ref{sec:conclusion} we summarize the progress achieved in the work.

\section{Related works}\label{sec:related_works}

Although the topic is extremely broad, four primary themes present in the related literature and research papers can be isolated from it. These topics include earthquake-related damage to structures, detection and management of this damage, the use of UAVs for this purpose, and the use of synthetic datasets for machine learning algorithms.

On the subject of damage to buildings caused by earthquakes, many scientific works focus on the analysis of specific cases of disasters~\cite{Eem2018,Erener2021}, or individual constructions~\cite{Maeda2018} affected by earthquake. Related to them are works on estimating possible future damage in a given area~\cite{Kahandawa2018} and building general models of damage caused by earthquakes~\cite{Lang2013,Amiri2018} or management strategies. There are also works describing the assessment of the accuracy of currently used methods of seismic measurements in relation to the damage recorded on buildings~\cite{Astroza2008}, or novel systems of seismic data collection~\cite{Stone2018}. 

The field of detection and management of identified faults is also rich in research. In the detection of defects, methods employing, among others, dynamic response of the structure~\cite{Shang2020} or laser scanning are used~\cite{Rabah2013}, but for a long time there has also been a significant increase in the number of works devoted to the use of computer vision, also regarding earthquakes~\cite{Rezaeian2010}. In this sub-field, classical methods of computer vision~\cite{Adhikari2012} are currently being replaced by methods that derive from machine learning, using convolutional~\cite{He2020} and fully convolutional~\cite{Mei2020} neural networks, LSTM~\cite{Zhang2018} networks or other techniques combining~\cite{Ni2019} or improving~\cite{CNNmod} upon these methods. However, it should be noted that only a few systems have been dedicated to detecting more than a single type of defect. Similarly, the field of identified damage management is still evolving, using, among other things, BIM models~\cite{Lin2016} and mobile applications~\cite{Sadi2014}. 

The use of UAVs is extremely closely related to the detection of damage to buildings, as they have been used for this purpose for a long time, not only in the form of flying vehicles, but also self-propelled rovers~\cite{Li2017}. Flying vehicles were used to detect damage on various surfaces, such as pavements~\cite{Zhang2019}, railroads~\cite{Wu2018} and public infrastructure facilities~\cite{Yang2018}, also with additional sensors~\cite{Addabbo}. 

The last field -- the use of synthetic data sets in training machine learning algorithms also has a long history of research related to it. Synthetically generated data sets do not necessarily have to be images and have been used in many areas, ranging from sociology~\cite{Burgard}, finance~\cite{Lopez2014}, medicine~\cite{Arvanitis}, to the issues related to computer vision. Since sets of correctly labeled data are necessary in the training of ML algorithms, and their manual collection is extremely time-consuming, automatic generators of synthetic data were also developed, thus further reducing the laboriousness of building a data sets~\cite{Mendonca}. 

\section{Our approach}
\label{sec:our_approach}
In our approach, we focused on putting various computer vision and machine learning techniques to the test to find the optimal solution to the problem at hand. We used various models of Convolutional Neural Networks for the task of image recognition and semantic segmentation, and decision trees, random forests and naive Bayes algorithm for the classification task. While developing our solution, we used ready-made state-of-the-art models with transfer learning technique for feature recognition as well as our own models and algorithms, developed exclusively for the task and trained from scratch.

In conclusion, we managed to develop a robust pipeline that can perform the tasks of segmenting construction components, their defects and assessing each of the elements' condition in the single run of the algorithm, while delivering satisfactory accuracy. It has to be noted, however, that the presented solution was validated only with the given, synthetic dataset, and for the practical usage, the models should be fine-tuned also on the real-world data points.

All the scripts needed to replicate the obtained results are provided in our repository~\cite{repository}. The experiments described in this paper were performed on machine equipped with Intel Core i7 3.80 GHz CPU, Nvidia GeForce RTX 3080Ti GPU and 64 GB of RAM.

\subsection{Dataset description and initial management}
\label{ssec:dataset_mgm}

\subsubsection*{Dataset description}

The provided dataset~\cite{ICDataset} consisted of $4\:808$ images of $1920 \times 1080px$ size from artificially generated drone flight in 3D urban surroundings and depicted multi-story apartment buildings with hardly noticeable defects along with similarly looking backgrounds. Out of all images, $3\:804$ were labeled into tasks of component, defect (cracking, spalling and exposed rebar), and damage state recognition. Additionally, separate labels were provided for depth channel of the images. 

However, we found the direct usage of the dataset proved to be difficult, as it came with a set of problems that had to be resolved in the first step of data preparation. The most important problem was the presence of conflicting labels in the task of defect recognition. The total number of images with colliding labels along with label-to-label collision marking is summarized in Tab.~\ref{tab:collisions}. This problem prevented the use of a single model for defect recognition, as it would have to yield the probabilities of occurrence of three separate labels. This problem was solved by training three separate models for each of the defect class.

\begin{table*}[h]
	\centering
	\begin{tabularx}{.95\linewidth}{p{.45\linewidth}c}
		\multirow{2}{*}{\textbf{Collision:}} & \multirow{2}{*}{\textbf{\begin{tabular}[c]{@{}c@{}}Number of images where \\ collisions occurred:\end{tabular}}} \\
		&                                                                                                       \\ \hline \hline
		Cracking - Exposed rebar             & 559                                                                                                   \\
		Cracking - Spalling                  & 3660                                                                                                  \\
		Exposed rebar - Spalling             & 1931                                                            
	\end{tabularx}\label{tab:collisions}
	\caption{Occurrences of label collisions in the dataset}
\end{table*}

Next, we discovered that the defect labels themselves make up only a small fraction of the whole labeled image, which was most noticeable for the \textit{exposed rebar} class. The pixels marked as exposed rebar occupied a maximum of only $1\%$ of the label image, with occurrences of as few as only one labeled pixel per image (less than one hundredth of a percent of all pixels in the image). Furthermore, it was found out that many of the images do not contain any of the class labels, again, with \textit{exposed rebar} class not present on nearly half of the training images. To mitigate this problem, we introduced a step-by-step approach, in which at first the building in image was split into its components and only then the defects were sought in the area of segmented out components. 

The last identified problem concerned the buildings visible in the background. While such image alignment is realistic, it does not coincide with the assigned labels, as they cover only the building in the foreground, and thus would be misleading for the classifier during its training (\emph{eg.} wall in the more distant building would be considered \textit{background} while similarly looking wall in the building nearby would be classified accordingly to its class). Moreover, the provided depth channel cannot be used directly by the models, as this data is not provided for the evaluation dataset (although RGB-D sensors are already in use with UAVs~\cite{Taozhang}). To avoid this problem along with the need to estimate depth maps with ML algorithms like in already existing works~\cite{Khan2020}, we decided to train a simple segmentation model to differentiate between background and foreground in the initial step of the image analysis.

\subsubsection*{Initial dataset management}

Before developing, training and testing our algorithms, a series of modifications to the dataset had to be done, accordingly to the issues described above. The first one was to split it into training and testing datasets to avoid overfitting of the trained algorithms and allow for reliable metrics checking. We split the dataset randomly in 4:1 ratio, where $20\%$ of the dataset was intended for testing, thus yielding $3\:043$ datapoints for training and $761$ for testing. To preserve the split and be able to repeat it with the same outcome, it was initiated with the known seed for the random algorithm.

\begin{figure*}[h!]
\centering
\includegraphics[width=1\textwidth]{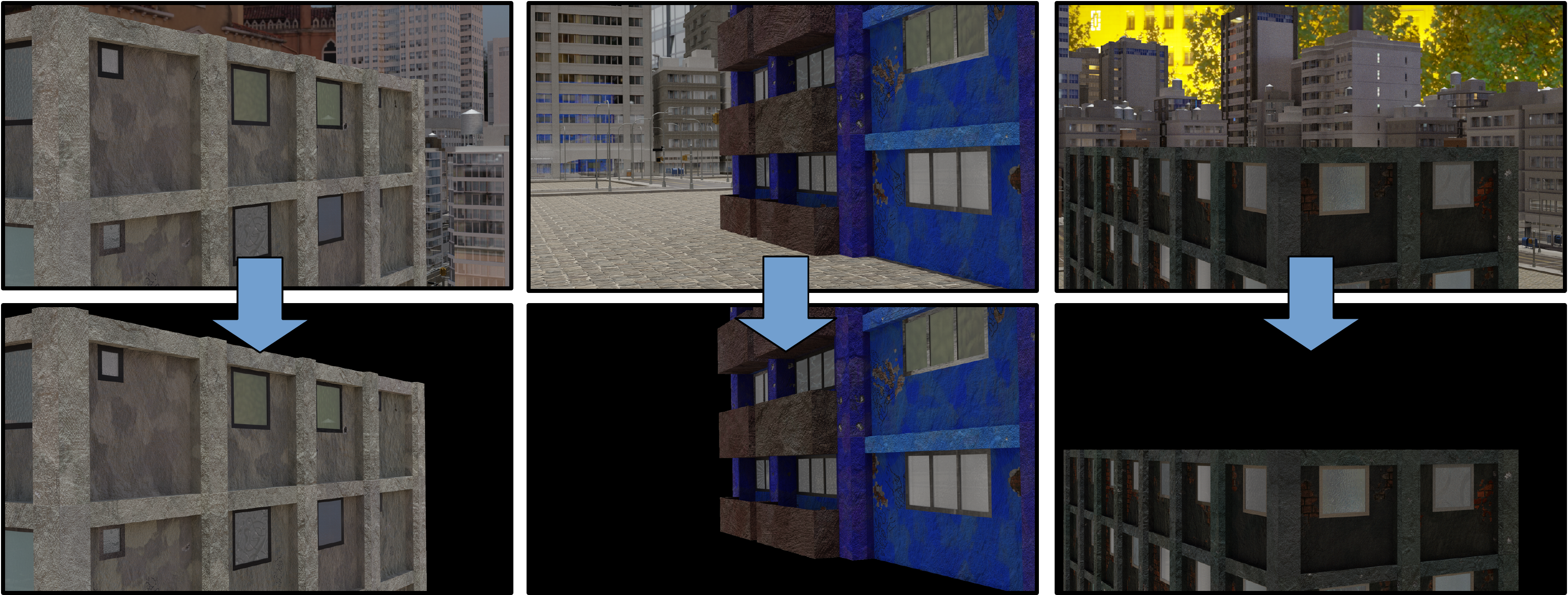}
\caption{Initial dataset images (top) and reworked images by background removal (bottom)\label{fig:background_rework}}
\end{figure*}

After splitting the dataset, we prepared it for the task of detecting objects in the foreground and masking unwanted background ones. To include this step in the single, continuous pipeline of performing all the given tasks in one algorithm run, we added it as \textit{Task~0}. To prepare the dataset, we provided simple, binary image mask labels for classes \textit{background} and \textit{foreground}, where all the objects from component segmentation task were counted as \textit{foreground} class. This way, the data prepared for classifier could differentiate only between whole, close and distant set of objects instead of each object distance type separately. The sample images from the dataset and their reworked counterparts are depicted in the Fig.~\ref{fig:background_rework}. Apart from \textit{Task~0}, images reworked for this dataset, albeit with different labels, would be then later used for the rest of the tasks as well.

Next, for the task of detecting defects, we prepared a new dataset that could help us with the problem of small number of labeled pixels for the defects and their complete absence from some of the images. To do so, we used labels for detecting components from \textit{Task~2} to make rectangular crops containing a single element and its immediate surroundings. We repeated the same operation on label images, thus obtaining a dataset containing a higher number of smaller than initial images, but to greater extent filled with labeled pixels. In this step, we also excluded all the image parts, where the defects were not visible, to slightly counteract biasing the classifiers towards \textit{background} class. The final yield of images for every class in both training and testing datasets for defect detection along with the total sum of damaged elements is summarized in Tab.~\ref{tab:split_images_num}.

\begin{table*}[h]
	\centering

	\vspace*{3pt}
	\begin{tabularx}{.95\linewidth}{Xccc}
		\textbf{Class} & \textbf{Training set} & \textbf{Testing set} & \textbf{\begin{tabular}[c]{@{}c@{}}Total damaged \\ components\end{tabular}} \\ \hline \hline
		Cracking       & 11771                 & 5072                 & \textbf{16843}                                                                             \\
		Spalling       & 15681                 & 5030                 & \textbf{20711}                                                                             \\
		Exposed rebar  & 1394                  & 355                  & \textbf{1749}                                                                            
	\end{tabularx}\label{tab:split_images_num}
	\caption{Data points extracted for defects classes after the dataset modification}
\end{table*}

For the component segmentation task, we mostly used the initial dataset provided with the task. The only modifications that we made were the use of images modified for the background-foreground segmentation and re-coding the labels' numbers to integers -- this way, as image labels we obtained images containing only a single channel.

\begin{figure*}[ht!]
\centering
\includegraphics[width=0.8\textwidth]{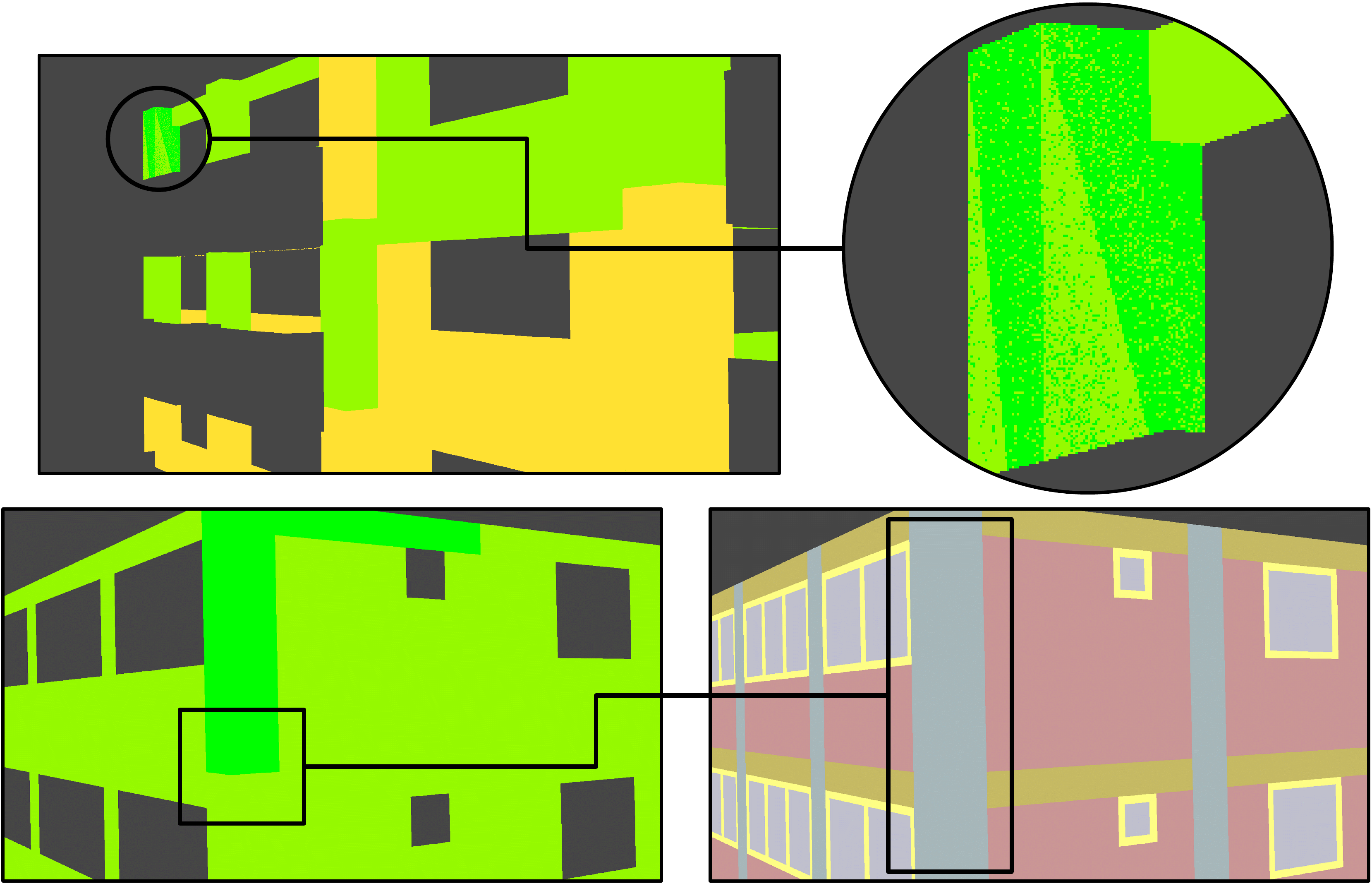}
\caption{Example of damage state marking problems: contrasting markings on a single element (top), no clear division of component segments (bottom)\label{fig:damage_state_problem}}
\end{figure*}

Lastly, we also modified the dataset for the last task -- the damage state assessment. However, this task turned out to be more demanding, as additional problems occurred. Some of the elements, while having their state decreased (by implication by the defects occurring on them), had no visible defects that were indicated by provided labels. Others had conflicting damage state designations even though the entire component should have been classified to a single state. Furthermore, the dataset containing structure elements did not separate their individual segments (for instance segmentation task), even though two segments of the same element may have differed in condition. The latter two of the problems are depicted in Fig.~\ref{fig:damage_state_problem}.

The problems described are beyond the ability to solve them using dataset manipulation or even an elaborate ML algorithm, and are a direct result of the way the dataset was constructed. As these were human errors made during the development of the dataset, possibly due to work fatigue, this draws all the more attention to the need for careful preparation of training data. However, we still attempted to solve them by approaching them in a more general way. Rather than focusing on elements' defects themselves, we decided to focus on their surfaces, and assess them by their general appearance.

\begin{figure}[h!]
\centering
\includegraphics[width=0.6\textwidth]{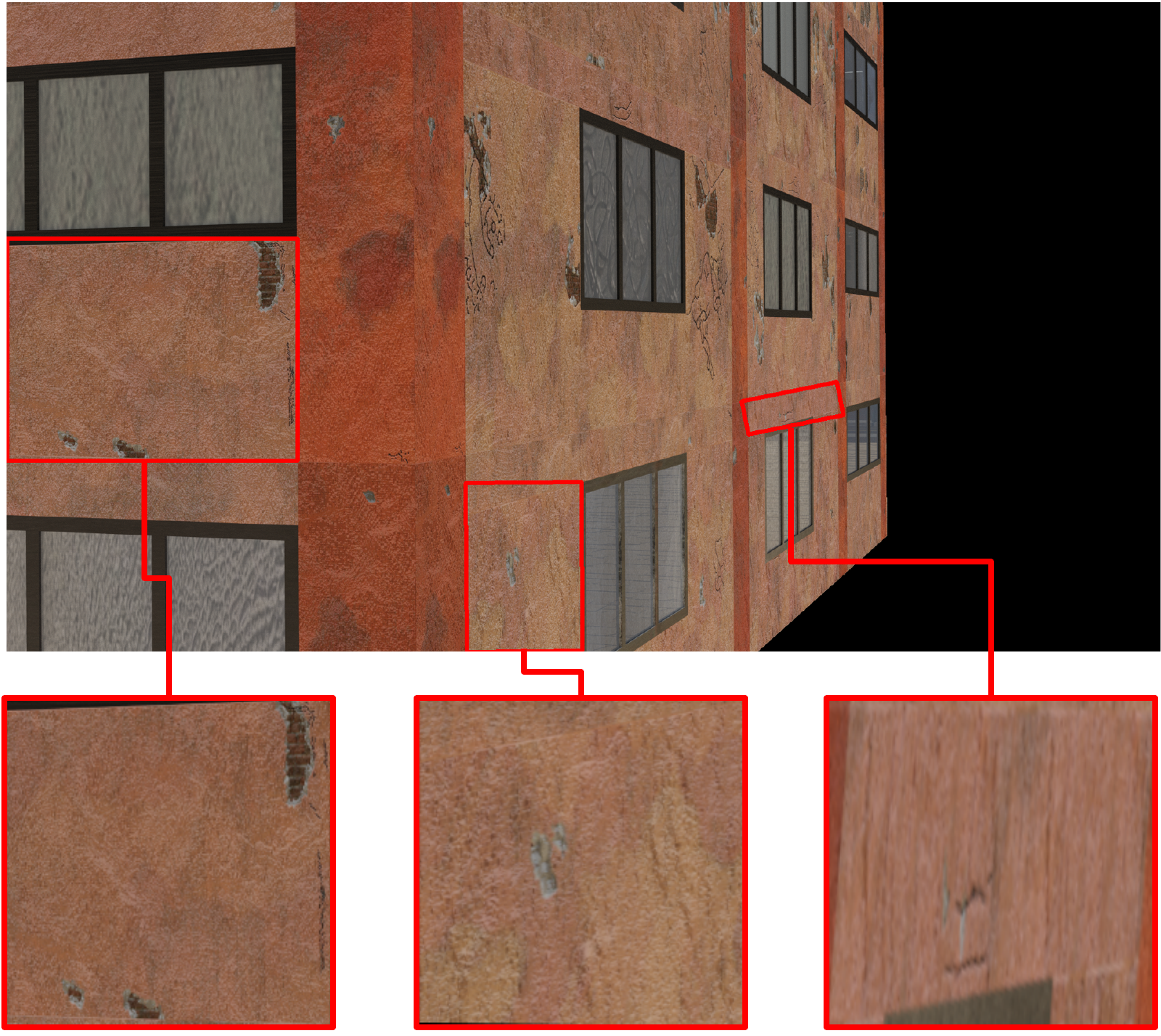}
\caption{Surface images warped to square shape\label{fig:imgs_warped}}
\end{figure}

To do so, we once again focused on extracting surface images from single elements. However,  this time, to retain as much information about the surface of a single element as there was, we extracted them with their minimum area, rotated rectangular bounding boxes and warped them into square shape of $ 224 \times 224px $. An example of such transformation is depicted in Fig.~\ref{fig:imgs_warped}. Such rectangular images were then used as an input to an image classifier, where the class was indicated by elements' damage state. The final yield of textures used for training is summarized in Tab.~\ref{tab:damage_state_split}. While considering the dataset, it is important to notice that in this task there are large inequalities between classes that, when mishandled, can affect the classifier.


\begin{table*}[h]
	\centering

	\vspace*{3pt}
	\begin{tabularx}{.95\linewidth}{Xccc}
		\textbf{\begin{tabular}[c]{@{}l@{}}Damage\\ state\end{tabular}} & \textbf{\begin{tabular}[c]{@{}c@{}}Training\\ dataset\end{tabular}} & \textbf{\begin{tabular}[c]{@{}c@{}}Testing\\ dataset\end{tabular}} & \textbf{\begin{tabular}[c]{@{}c@{}}Total extracted\\ components\end{tabular}} \\ \hline \hline
		No damage & 1025 & 254 & \textbf{1279} \\
		Light damage & 21811 & 5325 & \textbf{27136} \\
		Moderate damage & 35762 & 8825 & \textbf{44587} \\
		Severe damage & 2462 & 613 & \textbf{3075}
	\end{tabularx}\label{tab:damage_state_split}
	\caption{Datapoints extracted for damage state classes after the dataset modification}
\end{table*}

\subsubsection*{The impact of dataset management}

While the purpose of the modifications that we did to the dataset may be unclear at first, they have greatly assisted us in achieving high accuracy of the models we trained. Although in the next sections we'll be focusing on our most successful models developed with the data modified as described above, we will still provide the results from our other models that used the dataset directly or through simple extraction of data for additional comparison.

\subsection{Task 0 -- detecting foreground objects}
\label{ssec:task0}

As mentioned previously in Section~\ref{ssec:dataset_mgm}, we added the initial task marked as \textit{Task~0} to differentiate between objects in background and foreground more easily. To perform the segmentation, we used the previously prepared dataset along with model trained with Detectron 2 framework, taken from Detectron 2 model ZOO repository~\cite{detectron2}. Our choice of framework was based on the multiplicity of training options, choices of available architectures, and the ease of inference with the model. Detectron also allows for changing the type of task (\emph{eg.} object detection, semantic segmentation etc.) with only minimal changes to the code, for the ease of testing different approaches.

During our tests we found out that the model giving us optimal results considering both accuracy and inference speed was the Faster R-CNN~\cite{ren2016faster} based on ResNet 50 model~\cite{ResNet50} with additional fully convolutional head as feature proposal network, trained initially for 12 epochs on ImageNet~\cite{deng2009imagenet} dataset. We fine-tuned the model for $15$ additional epochs using learning rate of $2.5e-4$ with batches containing two images each and no additional learning rate decay.

The metrics of the model we obtained after training, including both -- mean values and values per class of IoU and accuracy per pixel, are presented in Tab.~\ref{tab:task0_metrics}. We found them satisfactory, as masks resulting from inference with the model hardly differ from the ones we manually prepared for training, and can be used in subsequent tasks without additional modifications.

\begin{table}[h]
	\centering
	\begin{tabularx}{.75\linewidth}{Xc}
		\textbf{Metric} & \textbf{Value {[}\%{]}} \\ \hline \hline
		mean IoU & 98.88 \\
		mean pixel accuracy & 99.28 \\
		IoU - foreground & 99.46 \\
		IoU - background & 98.30 \\
		\begin{tabular}[c]{@{}l@{}}Foreground pixel\\ accuracy\end{tabular} & 99.87 \\
		\begin{tabular}[c]{@{}l@{}}Background pixel\\ accuracy\end{tabular} & 98.70
	\end{tabularx}\label{tab:task0_metrics}
	\caption{Metrics of foreground detection model}
\end{table}

\subsection{Task 1 -- detecting defects}
\label{ssec:task1}

Similarly to \textit{Task~0}, \textit{Task~1} also focuses on image semantic segmentation with intention to detect cracks, spalling and exposed rebar on the surface of the structure. This time, however, the objects sought in the image are much smaller and more sparse across the dataset. As stated previously in Section~\ref{ssec:dataset_mgm}, we attempted to mitigate this problem by extracting only those parts of the dataset, where defects were present. Also, due to the collisions of labels described in the description of the dataset, we had to train one model per defect separately.

\begin{figure*}[h!]
\centering
\includegraphics[width=0.9\textwidth]{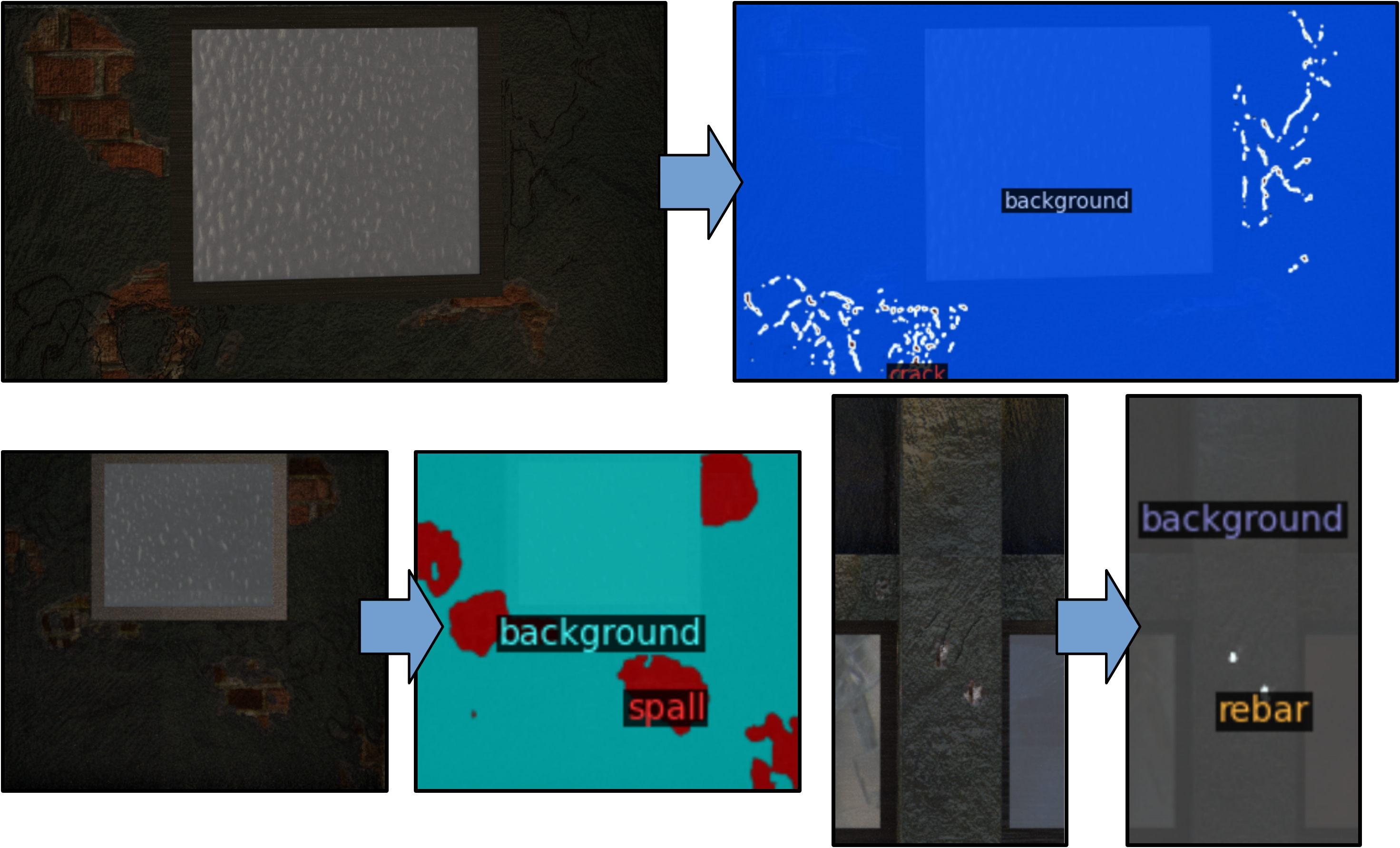}
\caption{Example predictions for each model: cracking (top), spalling (bottom left) and exposed rebar (bottom right)\label{fig:task1_preds}}
\end{figure*}

With such prepared dataset and task, we used the same architecture as in \textit{Task~0} for three models, as it again proved to provide the best balance between accuracy and inference time. To achieve the best results, we changed training parameters -- this time each model was fine-tuned for additional 50 epochs with learning rate decaying from initial $2.5e-4$ by $\frac{2.5e-4}{50}$ every epoch. The resulting metrics (IoU and pixel accuracy per defect) calculated on the testing dataset after training are presented in Tab.~\ref{tab:task1_metrics}, and sample images from model inference are depicted in Fig.~\ref{fig:task1_preds}. Tab.~\ref{tab:task1_metrics} also provides comparison between models trained with modified and unmodified dataset.

\begin{table*}[h]
	\centering

	\resizebox{0.70\textwidth}{!}{\begin{tabular}{l|ccc|ccc}
		\multirow{2}{*}{\textbf{Metric}} & \multicolumn{3}{c|}{\textbf{After dataset modification}} & \multicolumn{3}{c}{\textbf{Before dataset modification}} \\
		& Cracking & Spalling & \begin{tabular}[c]{@{}c@{}}Exposed\\ rebar\end{tabular} & Cracking & Spalling & \begin{tabular}[c]{@{}c@{}}Exposed\\ rebar\end{tabular} \\ \hline \hline
		Mean IoU {[}\%{]} & \textbf{64.16} & \textbf{89.42} & \textbf{64.86} & 63.83 & 88.30 & 51.81 \\
		Defect IoU {[}\%{]} & \textbf{29.73} & \textbf{80.12} & \textbf{29.81} & 28.57 & 77.45 & 3.62 \\
		Background IoU {[}\%{]} & 98.59 & 98.73 & 99.91 & \textbf{99.08} & \textbf{99.13} & \textbf{99.99} \\
		\begin{tabular}[c]{@{}l@{}}Mean \\ pixel accuracy {[}\%{]}\end{tabular} & \textbf{67.51} & \textbf{93.69} & \textbf{69.58} & 67.39 & 92.58 & 52.25 \\
		\begin{tabular}[c]{@{}l@{}}Defect \\ pixel accuracy {[}\%{]}\end{tabular} & \textbf{35.34} & \textbf{87.96} & \textbf{39.19} & 35.02 & 85.51 & 4.50 \\
		\begin{tabular}[c]{@{}l@{}}Background \\ pixel accuracy {[}\%{]}\end{tabular} & 99.68 & 99.43 & 99.97 & \textbf{99.76} & \textbf{99.64} & \textbf{99.99}
	\end{tabular}}\label{tab:task1_metrics}
	\caption{Metrics of defects detection models}
\end{table*}

As seen in Tab.~\ref{tab:task1_metrics}, our modifications to the dataset helped with the problem of model bias towards detecting \textit{background} class and made them more balanced, especially in \textit{Exposed rebar} task, where pixel accuracy increased over eightfold. In Fig.~\ref{fig:task1_preds} it can also be seen how small an area the searched defect could be, especially considering the initial, much greater size of the input image, what once again justifies the best result achieved by the largest type of defect (spalling). The results however still are not perfect, and other means such as random undersampling of dominant class or training performed for image patches could also be considered.

\subsection{Task 2 -- segmenting out components}
\label{ssec:task2}

\textit{Task~2} once again focuses on image semantic segmentation, so we used methods similar to \textit{Task~1} and~\textit{2}. We used Detectron 2 framework to fine tune ResNet 50 with Feature Proposal Network on dataset with background objects removed. We achieved satisfactory results when the model was trained for additional $ 15 $ epochs and learning rate an order of magnitude greater than for the previous tasks -- $ 2.5e-3 $. We also did not use learning rate decay schedule this time. 

\begin{figure*}[h!]
\centering
\includegraphics[width=0.9\textwidth]{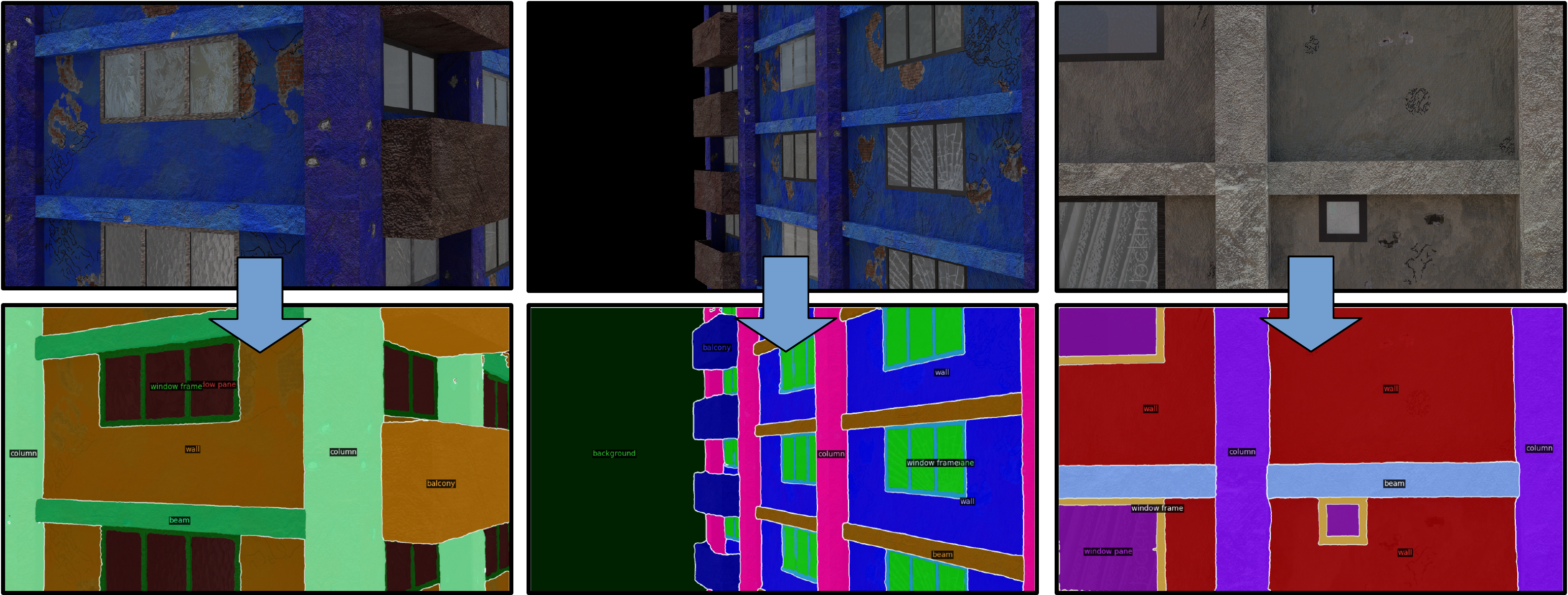}
\caption{Example results obtained after performing initial background removal\label{fig:task2_res1}}
\end{figure*}

\begin{figure*}[h!]
\centering
\includegraphics[width=0.9\textwidth]{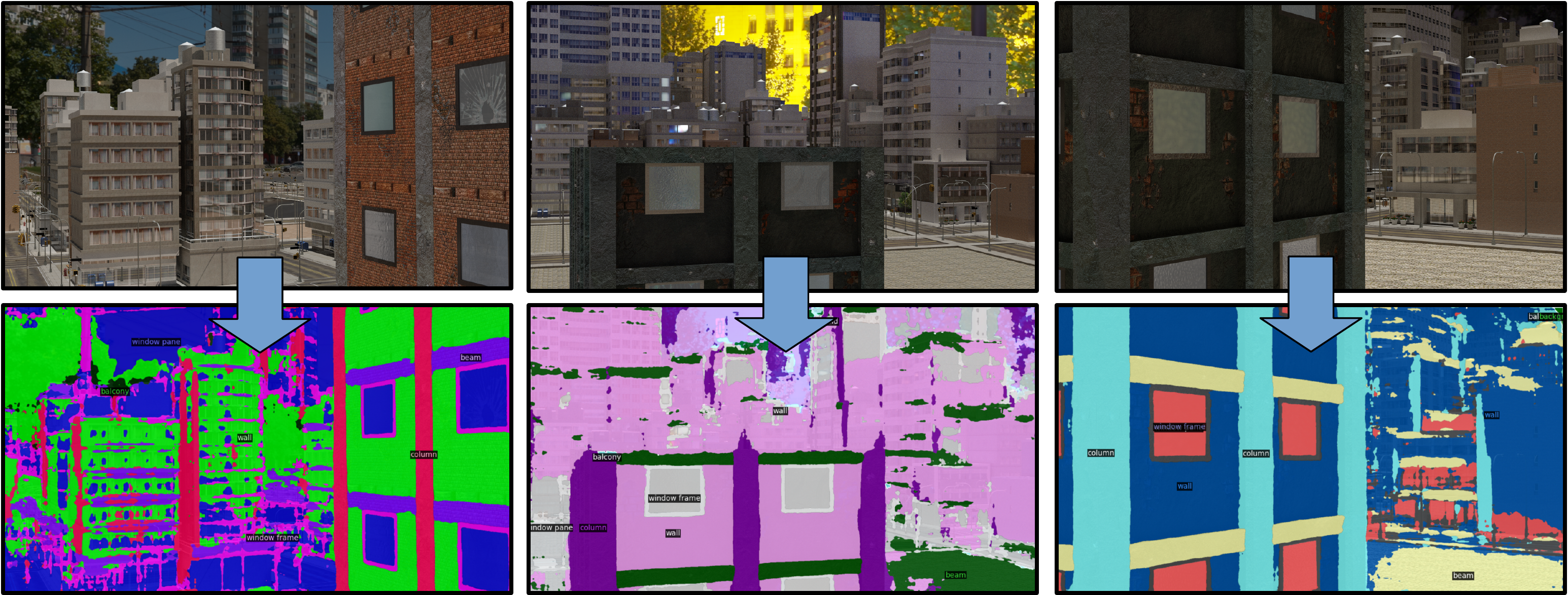}
\caption{Example results without background removal\label{fig:task2_res2}}
\end{figure*}

The resulting metrics we obtained after the training for both datasets -- the initial one and after background removal with model from \textit{Task~0} -- are summarized in Tab.~\ref{tab:task2_metrics}. Moreover, Figs.~\ref{fig:task2_res1} and~\ref{fig:task2_res2} show the example results for component detection for both cases -- when background is and is not removed.

\begin{table*}[h!]
	\centering
	\caption{Metrics of component segmentation model}
	\vspace*{3pt}
	\resizebox{0.90\textwidth}{!}{
	\begin{tabular}{l|cccccccc|c}
        Metric & Wall & Beam & Column & \begin{tabular}[c]{@{}c@{}}Window\\ frame\end{tabular} & \begin{tabular}[c]{@{}c@{}}Window\\ pane\end{tabular} & Balcony & Slab & Background & Mean \\ \cline{2-10} 
         & \multicolumn{8}{c|}{\textbf{After background removal}} &  \\
        IoU {[}\%{]} & \textbf{91.42} & \textbf{89.63} & \textbf{84.25} & \textbf{82.14} & \textbf{97.31} & \textbf{96.84} & \textbf{90.81} & \textbf{99.89} & \textbf{91.54} \\
        Pixel accuracy {[}\%{]} & \textbf{98.81} & \textbf{93.73} & \textbf{86.27} & 87.38 & 97.86 & \textbf{98.64} & \textbf{94.51} & \textbf{99.96} & \textbf{94.65} \\ \hline
         & \multicolumn{8}{c|}{\textbf{Without background removal}} &  \\
        IoU {[}\%{]} & 70.53 & 57.31 & 69.69 & 59.35 & 76.31 & 87.47 & 0.82 & 0.35 & 52.73 \\
        Pixel accuracy {[}\%{]} & 98.70 & 92.79 & 86.05 & \textbf{87.48} & \textbf{98.06} & 97.88 & 0.86 & 0.35 & 70.27
    \end{tabular}
	}\label{tab:task2_metrics}
\end{table*}

As can be seen in Fig.~\ref{fig:task2_res1}, \ref{fig:task2_res2} and Tab.~\ref{tab:task2_metrics}, removing the background has significantly improved the performance of the model, especially in the case of detecting slabs and backgrounds, whereas the accuracy with visible background was negligible. The only elements whose pixel-by-pixel detection accuracy slightly increased without background removal were parts of windows, and this may have been due to their specific shape in the dataset, which windows found in buildings in the background did not have.

At the same time, it should be noted that the model working on images without first removing the background still correctly recognizes the elements of objects in the background to some extent, even though they differ significantly in appearance from those in the dataset. This indicates, on the one hand, that the model was correctly chosen for the task, since it was able to generalize the acquired knowledge beyond the provided dataset, and on the other hand, the necessity to apply the first step of the analysis -- removing the background, since the results of the actually conducted measurements would be muddied by the occurrence of false-positive detections.

\subsection{Task 3 -- assessing the damage state}
\label{ssec:task3}

In the first attempt to solve \textit{Task~3} we intended to use the relationship between an element, its total area, and the relative area of each defect as a one-dimensional input vector for a shallow machine learning algorithm. We used decision trees with maximal depth of $59$ splits, random forest with total of $200$ models and naive Bayesian algorithm with and without data normalization. The training data we collected came from the partitioning of the structure into elements by the model trained for \textit{Task~2} and the damage detected using the model from \textit{Task~1}. This way we managed to gather $61\:058$ data points for training and $15\:036$ for testing. The final vector along with its exemplary data presented itself in this way:

		\begin{equation}
		\left\{ E_t, E_{sr}, C_r, R_r, S_r \right\} = E_s
		\end{equation}
		
		
		\begin{math}
		\emph{eg.} \left\{ 3.0, 0.0245, 5.90e-5, 3.94e-5, 9.60e-3 \right\} = 3.0
		\end{math}
where

	\begin{math}
	where: \\ E_t: element\;type\;(mapped\;to\;int\;value) \\ E_{sr}: element\;size\;to\;image\;size\;ratio \\ C_r: crack\;size\;to\;element\;size\;ratio \\ R_r: exposed\;rebar\;size\;to\;element\;size\;ratio \\ S_r: spalling\;size\;to\;element\;size\;ratio \\ E_s: element\;damage\;state \vspace*{3pt}
	\end{math}

However, despite a vast dataset for model development, this approach proved to be inaccurate. It was mostly due to the component defect not visible in the frame of the image and occasional false negative indications of the model from \textit{Task~1}. This led to many instances of data points where even though the element had no visible defect, its state was described as damaged. The results of models trained using this approach are shown in Tab.~\ref{tab:task3_metrics}. 

\begin{figure*}[h!]
\centering
\includegraphics[width=\textwidth]{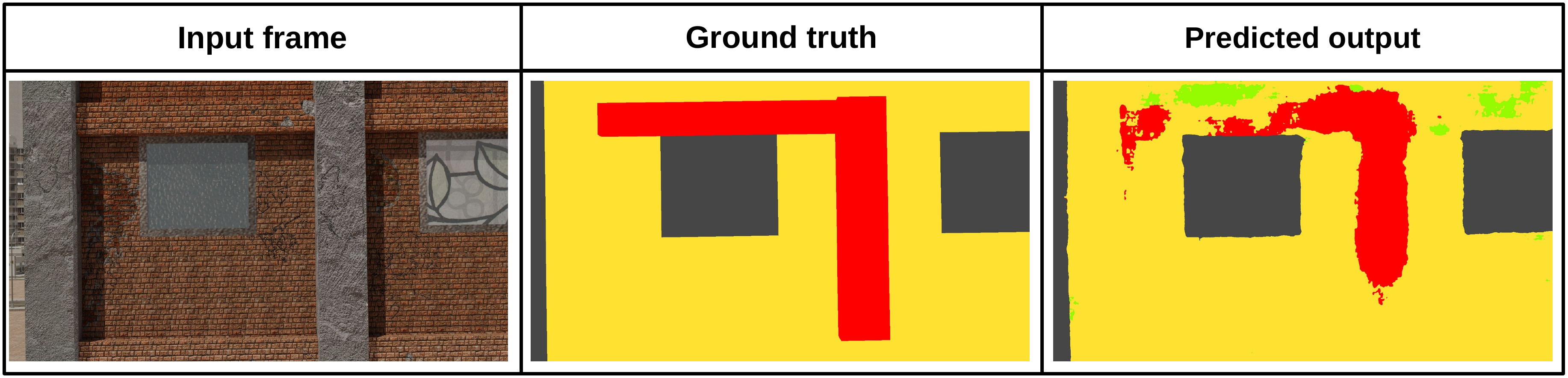}
\caption{Single model approach compared to ground truth label\label{fig:single_model}}
\end{figure*}

To increase the accuracy of classification, we turned to methods based on a single model that uses computer vision to determine the damage state of elements. We called this approach \textit{single model baseline}, where we used Detectron 2 directly on the visual data. Unfortunately, this approach turned out to be ineffective, not achieving the best result in any of the categories, as shown in Tab.~\ref{tab:task3_metrics}. In Fig.~\ref{fig:single_model}, we also show the result of the model inference compared to the label.

To improve damage state assessment, as stated previously in Section~\ref{ssec:dataset_mgm}, we chose the method that is approaching the problem in a more general way -- by image recognition rather than inference based on defects found by other models. Again, the part of the task related to the semantic segmentation of the structure components has been done before, and we can continue on the segmentation done with high accuracy in \textit{Task~2} to focus on image recognition task.

In order to perform the image recognition we used KrakN framework~\cite{ZARSKI2021} to train $4$ separate CNN models using transfer learning technique with 3-fold cross validation on dataset containing warped images of elements' surfaces. Models we used as feature extractors for single layer logistic regression classifier were: VGG16~\cite{simonyan2015deep}, Densenet~\cite{huang2018densely}, ResNet and Xception~\cite{chollet2017xception}, all of which were pre-trained on the ImageNet dataset. The results of the training are summarized in Tab.~\ref{tab:task3_metrics} -- note that this time the only measured metrics were accuracy per class, average accuracy and average F1 score which describes balance between precision and recall. This time IoU or pixel accuracy will be derived from the segmentation performed in \textit{Task 2} and have no impact on the overall score. 

\begin{table*}
	\centering

	\vspace*{3pt}
	\resizebox{0.80\textwidth}{!}{\begin{tabular}{lcccccc}
		\multicolumn{1}{c}{\multirow{2}{*}{\textbf{Model}}} & \multicolumn{4}{c}{\textbf{Accuracy per class {[}\%{]}}} & \multirow{2}{*}{\textbf{\begin{tabular}[c]{@{}c@{}}Average \\ accuracy {[}\%{]}\end{tabular}}} & \multirow{2}{*}{\textbf{\begin{tabular}[c]{@{}c@{}}Average\\ F1 score {[}\%{]}\end{tabular}}} \\
		\multicolumn{1}{c}{} & \textbf{\begin{tabular}[c]{@{}c@{}}No \\ damage\end{tabular}} & \textbf{\begin{tabular}[c]{@{}c@{}}Light \\ damage\end{tabular}} & \textbf{\begin{tabular}[c]{@{}c@{}}Moderate \\ damage\end{tabular}} & \textbf{\begin{tabular}[c]{@{}c@{}}Severe \\ damage\end{tabular}} &  &  \\ \hline \hline
		VGG16 & 28.74 & 74.25 & 83.13 & 59.11 & 78.06 & 77.83 \\
		Densenet & 28.35 & 78.14 & 86.33 & 61.17 & 81.40 & \textbf{81.13} \\
		ResNet & 28.34 & 77.56 & \textbf{86.93} & 58.30 & \textbf{81.42} & 81.09 \\
		Xception & 19.29 & 75.42 & 85.44 & 57.21 & 79.59 & 79.18 \\ \hline
		Decision tree & 12.89 & 70.61 & 76.61 & \textbf{61.96} & 72.98 & 73.19 \\
		Random forest & 0.00 & 81.55 & 77.50 & 38.36 & 76.00 & 75.29 \\
		Naive Bayes & \textbf{98.05} & 21.36 & 39.31 & 52.65 & 34.45 & 43.66 \\
		\begin{tabular}[c]{@{}l@{}}Naive Bayes\\ normed\end{tabular} & 0.00 & \textbf{98.62} & 13.97 & 28.73 & 35.73 & 35.87 \\
		\hline
		\begin{tabular}[c]{@{}l@{}}\textit{Single model}\\ \textit{baseline}\end{tabular} & \textit{23.68} & \textit{68.66} & \textit{80.40} & \textit{53.02} & \textit{64.67} & \textit{69.12}
	\end{tabular}}\label{tab:task3_metrics}

	\caption{Metrics of damage state detection model}
\end{table*}

As seen in Tab.~\ref{tab:task3_metrics}, while shallow machine learning techniques can provide near perfect accuracy for single defect types, they tend to be highly biased and thus resulting in low average accuracy and F1 score. On the other hand, deep CNN models provide more balanced results across classes and do not tend to over-promote particular predictions. At the same time, once again, it is important to account for highly imbalanced dataset. However, even with properly performed random undersampling with the best performing models (ResNet and Densenet), we did not manage to achieve better results.

It is also possible to consider using model ensembles with only slightly biased members and use a voting algorithm to pick the most probable prediction.

\subsection{The final pipeline}
\label{ssec:pipeline_fin}

\begin{figure*}[h!]
\centering
\includegraphics[width=\textwidth]{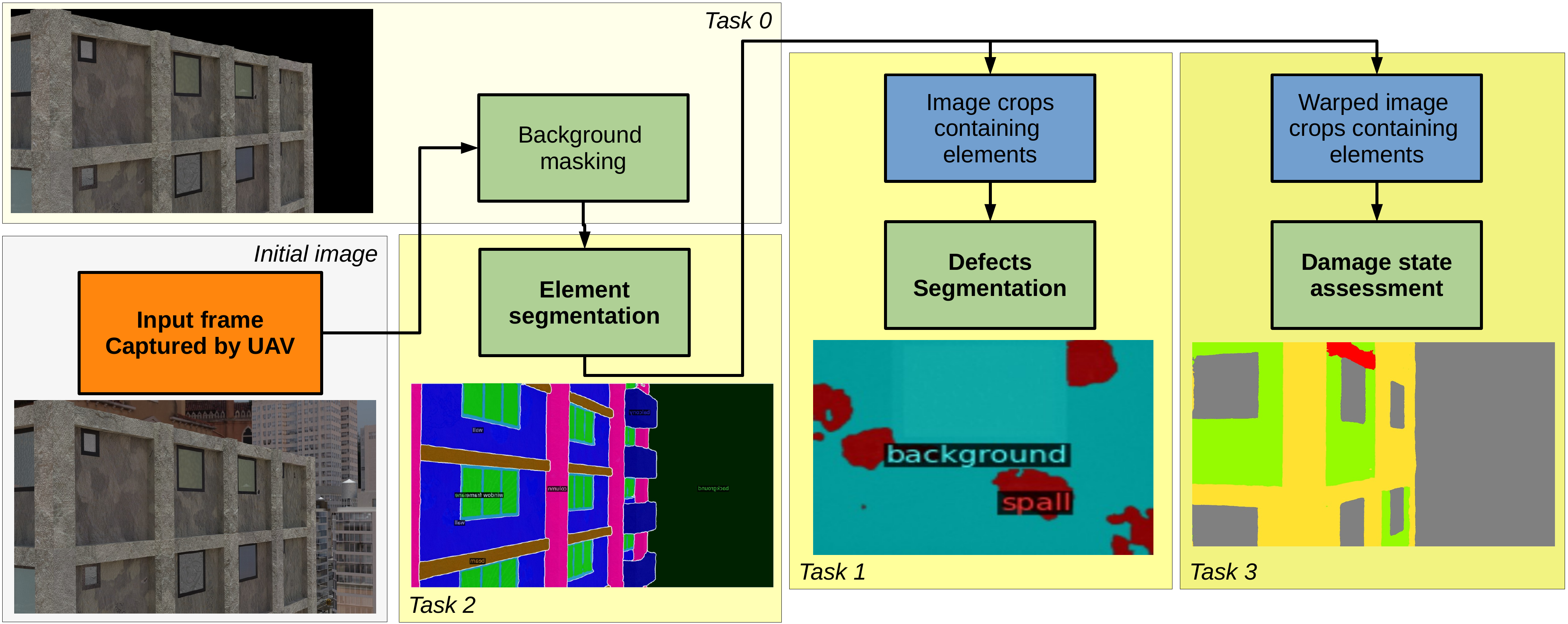}
\caption{The final pipeline of our solution\label{fig:final_pipeline}}
\end{figure*}

In Fig.~\ref{fig:final_pipeline} we present the final pipeline of our solution. 

First, the image retrieved by the UAV is stripped of background objects using the model trained for \textit{Task 0}. Then, the model from \textit{Task 2} performs structure segmentation to its components. These components after additional processing are used in further tasks. For \textit{Task 1}, the segmented parts of the structure come as rectangular frames of the image, and then damage is detected on them using the model for semantic segmentation. For \textit{Task 2}, they are further warped to a square shape, after which the CNN model performs an image recognition task on them.

While the tasks performed by the pipeline are backwards dependent, the great advantage of this approach is to divide one large task into a number of smaller tasks that can be performed with greater accuracy. At the same time, this approach allows for wider freedom in the manipulation of the models that make up the workflow -- models can be freely exchanged or fine-tuned with new batches of data as single nodes responsible for individual tasks, thus maintaining the integrity of the solution.

\section{Discussion}
\label{sec:discussion}

Considering the phasing of successive tasks performed by with our pipeline from Section~\ref{ssec:pipeline_fin}, we can expect marginally lower accuracy of the whole workflow, albeit thanks to nearly perfect accuracy of the first stages, the reduction will not be significant. Also, some of the models we presented in the pipeline can certainly be improved in the terms of their particular accuracy. However, due to the specific nature of the dataset, much higher prediction accuracy cannot be expected. 

Still, there are methods that can aid the accuracy of the algorithm. One is the previously mentioned use of small image patches to train a model that performs damage detection. This way the problem of highly imbalanced dataset can be avoided, and models would not be as much biased towards \textit{background} class. This problem can also be mitigated with the use of random oversampling techniques that could help by expanding underrepresented classes or with the use of weighted loss function -- focal loss~\cite{lin2018focal}. However, while the latter could help with the training process (although during the performed tests, the gain was insignificant), the former could lead to model biased towards a specific shape of defect of single class.

The last method that can have a positive impact on the accuracy of the algorithm is to transform the flat representation of objects in the image into a 3D point cloud. This way, the problem of defects not visible on an element in a single frame may no longer be relevant, since the defect detection model would operate on the entire 3D object. The transformation could be performed with photogrammetric methods employing RGB-D sensors like in \emph{eg.}~\cite{zollhofer2018state}. However, this approach also has some drawbacks. First of them is the need of depth data associated with RGB images, that was not available for the task. Second is the necessity of heavy computations performed during preprocessing of the data, excluding this approach from on-site usage. Additionally, machine learning models trained in 3D point cloud environment require significantly more computing power than those operating on RGB images.

\section{Conclusions}
\label{sec:conclusion}

The detection of earthquake caused defects to buildings is an extremely important issue, affecting both the safety of building occupants and the efficient management and restoration of the affected area. The use of UAVs for this purpose is an important step in the desired direction by which the effectiveness of immediate post-earthquake response can be increased. Moreover, fully autonomous vehicles armed with algorithms that enable automatic and immediate assessment of structure technical condition will allow more efficient management of emergency services human resources and labor intensity reduction of the whole process. However, for that purpose, vast datasets and robust machine learning models are needed. 

While real-world datasets like PEER Hub ImageNet~\cite{PEERImagenet} or Mexico City~\cite{MexicoDataset} earthquake datasets are available, they focus on high-magnitude earthquakes, which is beyond the scope of continuous monitoring of structures after minor earthquakes. Other datasets that contain individual defect types, on the other hand, do not consider a holistic approach to the structure monitoring problem. For this reason, until actual imaging data is collected from low magnitude earthquakes in highly urbanized areas, synthetic datasets are one of the best ways to pretrain neural algorithms.

In this paper, we proposed a workflow along with dataset transformations and models trained for the tasks of foreground objects, defects, and structure components detection, as well as image recognition methods for elements' damage state assessment. In the work we presented, we used a synthetic dataset that can serve as a base for training machine learning models and achieved satisfactory results in all four tasks.

During our work, we also identified some recurring issues with synthetic datasets that can significantly affect the trained models. Although synthetic data, especially those that simulate extremely rare and large-scale events, are essential in building training datasets, they must be free of errors in the objects represented and the classes given. At the same time, even a flawlessly prepared dataset cannot be the only source of information for a machine learning algorithm -- for this, one must also be exposed to real data in the final training phase. With this in mind, a worthwhile concept to consider while creating a synthetic dataset is instead of a single, homogeneous dataset, creating multiple smaller but more diverse datasets for which the real-world data would always fit closer within the spectrum of variants. 

Our work has also pointed possible directions for further development of the proposed methods. As we mentioned in Section~\ref{sec:discussion}, a next step with a holistic approach to assessing the condition of a building would be a much-needed improvement to the overall process of structure maintenance. However, it requires creating a three-dimensional model of the building that is faithful to the original, using RGB-D cameras, photogrammetry methods or multi-scale approach. At the same time, the effort should be put into developing diverse synthetic datasets along with their real-life counterparts.

\paragraph{Acknowlegements} This work was supported by the \textit{European Union} through the \textit{European Social Fund} as a part of a \textit{Silesian University of Technology as a Centre of Modern Education based on research and innovation} project, number of grant agreement: \textbf{POWR.03.05.00 00.z098/17-00} (B.W. and M.Ż.).


\end{document}